\pdfoutput=1
\documentclass[11pt]{article}

\usepackage[preprint]{acl}

\usepackage{times}
\usepackage{latexsym}

\usepackage[T1]{fontenc}

\usepackage[utf8]{inputenc}

\usepackage{microtype}

\usepackage{inconsolata}

\usepackage{graphicx}

\usepackage{amsmath,amssymb}
\usepackage{booktabs,tabularx,array,pifont,colortbl}
\usepackage{enumitem}

\makeatletter
\renewcommand\@makefntext[1]{\noindent\@makefnmark#1}
\makeatother
\makeatletter
\g@addto@macro\normalsize{%
  \setlength\abovedisplayskip{5pt plus 2pt minus 2pt}%
  \setlength\belowdisplayskip{5pt plus 2pt minus 2pt}%
  \setlength\abovedisplayshortskip{2pt plus 2pt}%
  \setlength\belowdisplayshortskip{5pt plus 2pt minus 2pt}%
}
\makeatother

\title{LatComm: Reusable Context Compression for Latent Communication between Large Language Models}

\author{
 \textbf{Huawei Lin\textsuperscript{1}},
 \textbf{Jinting Liu\textsuperscript{1}},
 \textbf{Tan Yu\textsuperscript{2}},
 \textbf{Weijie Zhao\textsuperscript{1}}
\\
\\
 \textsuperscript{1}Rochester Institute of Technology,
 \textsuperscript{2}NVIDIA
}

\begin{document}

\maketitle

\begin{abstract}
Systems that combine several large language models (LLMs) often use one context for many questions. When the context is passed between models as text, the receiving model processes all of its tokens again for every question. Avoiding this cost requires a context representation that is compact, reusable across questions, and readable by a different LLM. Existing compression and latent communication methods each provide only some of these properties. We introduce \texttt{LatComm}, a \textbf{Lat}ent \textbf{Comm}unication interface that provides all three by separating encoding from reading: the context is encoded once without the question, and the receiver selects what to read for each question. A frozen context encoder LLM compresses the context into a fixed-size packet and sends it to a different frozen receiver LLM, which reads the packet through cross-attention, with gates computed from its prompt. On 12 multiple-choice question-answering datasets, \texttt{LatComm} improves average accuracy over single-vector baselines by up to 10.2 points. On QuALITY, it reduces the receiver's input by $57\times$ and answers the questions of an article $3\times$ faster.
\end{abstract}

\section{Introduction}

\begin{figure}[t]
    \centering
    \includegraphics[width=\linewidth]{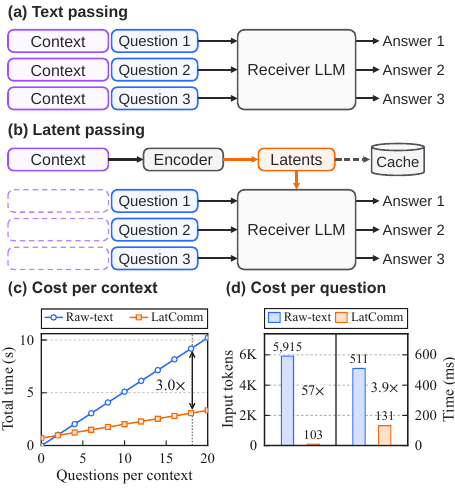}
    \vspace{-.25in}
    \caption{Text passing vs.\ latent passing. (a) The receiver reads the context again for every question. (b) The context is encoded once and reused. (c, d) Measured cost on QuALITY.}
    \label{fig:latcomm-teaser}
    \vspace{-.18in}
\end{figure}

\noindent\textbf{Motivation.}
Communication between large language models (LLMs) enables models to contribute different roles to a shared task, such as processing documents and answering questions \citep{wu2023autogen,hong2024metagpt,du2023debate,wang2025moa,lin2025agentomni,lin2026museautoskill}. In context-dependent workflows \citep{lewis2020rag,DBLP:conf/iclr/YaoZYDSN023}, the same context is often used many times: one document can support many questions, and one observation can inform later decisions. When the context is passed as text, the model that receives it, called the \emph{receiver}, processes all of its tokens again for every question (Figure~\ref{fig:latcomm-teaser}a). Shared contexts therefore lead to repeated computation across models and questions \citep{gim2024promptcache,liu2024cachegen,liu2024droidspeak}.
For example, in the reading comprehension dataset QuALITY \citep{pang-etal-2022-quality}, each article comes with 18.1 questions on average. The receiver reads 5,915 tokens for each question, and 98\% of them are the article (Figure~\ref{fig:latcomm-teaser}d).

\begin{table*}[!t]
\vspace{-.1in}
\centering
\begingroup
\small
\setlength{\tabcolsep}{2pt}
\renewcommand{\arraystretch}{1.1}
\definecolor{lcGreen}{RGB}{28,140,51}
\definecolor{lcRed}{RGB}{196,40,40}
\newcommand{\lcYes}{\textcolor{lcGreen}{\ding{51}}}
\newcommand{\lcNo}{\textcolor{lcRed}{\ding{55}}}
\newcommand{\lcHead}[1]{{\renewcommand{\arraystretch}{0.9}\begin{tabular}[c]{@{}c@{}}#1\end{tabular}}}
\begin{tabularx}{\linewidth}{@{}p{0.38\linewidth}>{\hsize=0.93\hsize\centering\arraybackslash}X>{\hsize=1.00\hsize\centering\arraybackslash}X>{\hsize=1.18\hsize\centering\arraybackslash}X>{\hsize=0.89\hsize\centering\arraybackslash}X@{}}
\toprule
\midrule
Method & \lcHead{Query-\\independent} & \lcHead{Fixed-budget\\compression} & \lcHead{Cross-LLM\\communication} & \lcHead{Bridge-only\\training} \\
\midrule
\mbox{LLMLingua-2~\citep{pan2024llmlingua2}} & \lcYes & \lcYes & \lcNo & \lcNo \\
\mbox{AutoCompressor~\citep{chevalier2023autocompressor}} & \lcYes & \lcNo & \lcNo & \lcNo \\
\mbox{ICAE~\citep{ge2024icae}} & \lcYes & \lcYes & \lcNo & \lcNo \\
\mbox{xRAG~\citep{cheng2024xrag}} & \lcYes & \lcYes & \lcNo & \lcYes \\
\mbox{CEPE~\citep{yen2024cepe}} & \lcYes & \lcNo & \lcNo & \lcNo \\
\mbox{DroidSpeak~\citep{liu2024droidspeak}} & \lcYes & \lcNo & \lcNo & \lcNo \\
\mbox{C2C~\citep{fu2025c2c}} & \lcNo & \lcNo & \lcYes & \lcYes \\
\mbox{LatentMAS~\citep{zou2025latentmas}} & \lcNo & \lcNo & \lcNo & \lcNo \\
\midrule
\textbf{\texttt{LatComm} (ours)} & \lcYes & \lcYes & \lcYes & \lcYes \\
\midrule
\bottomrule
\end{tabularx}
\endgroup
\vspace{-.1in}
\caption{Comparison of context compression and latent communication. Query-independent (reusable): the representation is produced without the question. Fixed-budget compression (compact): its size does not depend on the input length. Cross-LLM communication: latent representations pass from one LLM to a different LLM. Bridge-only training: only an interface between two frozen LLMs is trained.}
\label{tab:related-methods}
\vspace{-.2in}
\end{table*}

\noindent\textbf{Latent passing.}
Latent passing avoids this repetition (Figure~\ref{fig:latcomm-teaser}b). An encoder reads the context once and turns it into vectors, called \emph{latents}. The latents are stored and reused, so the receiver reads only the question. To replace the text, the latents need three properties. They must be \emph{reusable}: they are computed once, before any question is known. They must be \emph{compact}: their size does not grow with the context length, so the cost of storing and reading them stays fixed. They must be \emph{readable by a different LLM}, because systems often combine models of different sizes \citep{li2025edgecloud,belcak2025slm} and from different families \citep{chen2024frugalgpt,ong2025routellm,wang2025moa}. These properties are hard to combine. Latents built without the question and with a fixed size cannot keep only what one question needs, so the receiver must select the relevant information when it reads.

\noindent\textbf{Existing approaches.}
Existing methods provide only some of these properties. A receiver can store its key--value cache to avoid re-reading \citep{gim2024promptcache,liu2024cachegen}, but this cache grows with the context length and can be used only by the model that produced it. Context compression methods shorten what a model reads \citep{pan2024llmlingua2,mu2023gist,chevalier2023autocompressor}. ICAE compresses a context into fixed memory slots without the question, but it adapts the LLM with LoRA, and the slots are read by the same LLM \citep{ge2024icae}. xRAG keeps both models frozen and stores one vector per document \citep{cheng2024xrag}. This vector is the output of a pretrained retrieval encoder, so the amount of context that it carries is fixed by that encoder. A larger representation would need a larger projector, and its trainable parameters would grow with the size of the representation. Latent communication methods pass representations between models in place of text \citep{ramesh2025activations,DBLP:conf/acl/DuWBCZCZCY26,DBLP:journals/corr/abs-2511-03945,shi2025kvcomm}. C2C transfers representations between different frozen LLMs, but both models read the same input, so the transfer depends on the question and grows with the input length \citep{fu2025c2c}. Table~\ref{tab:related-methods} compares representative methods on four properties, and only \texttt{LatComm} has all four.

\begin{figure*}[!t]
    \centering
    \vspace{-.1in}
    \includegraphics[width=.96\textwidth]{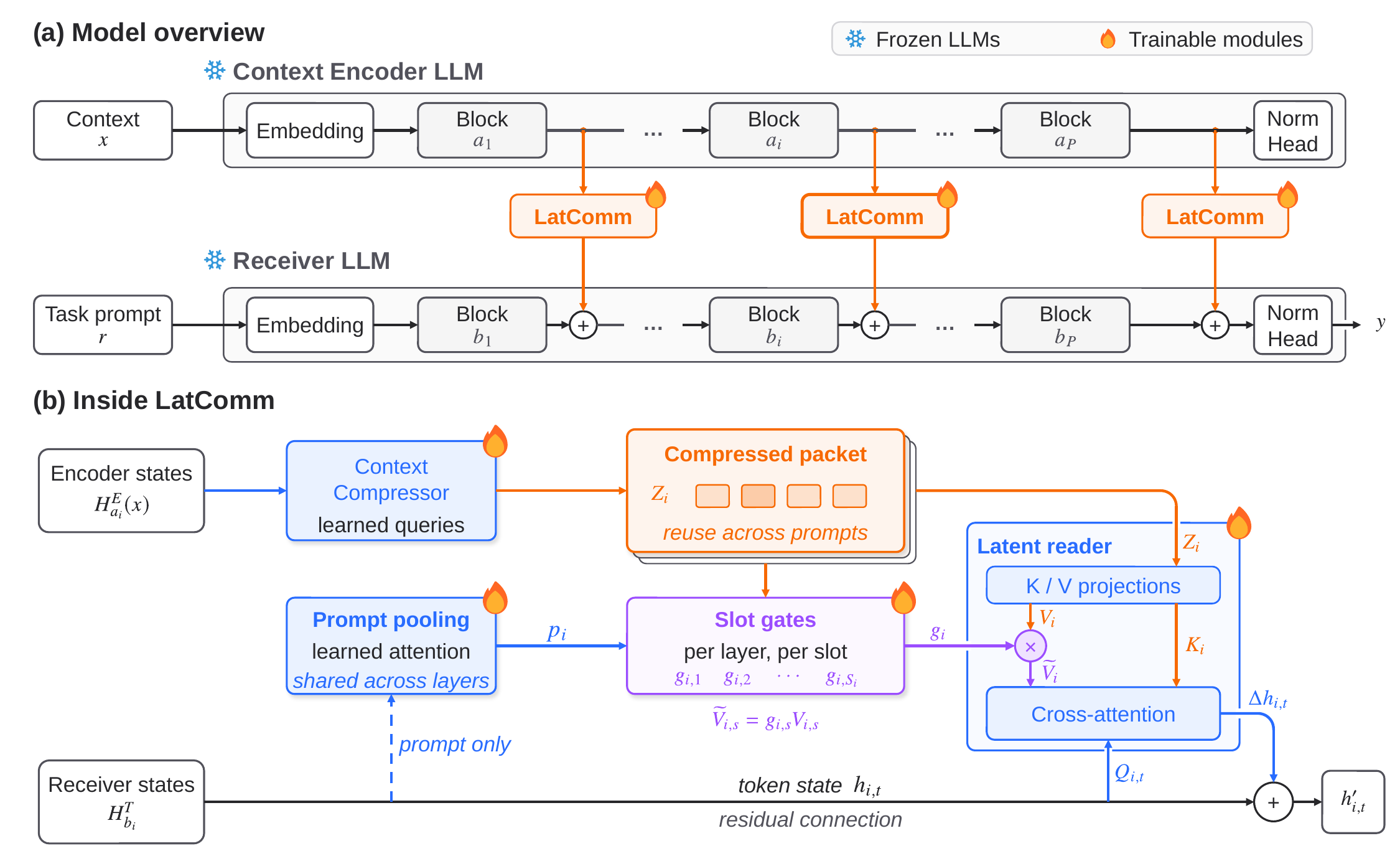}
    \vspace{-.1in}
    \caption{\texttt{LatComm} overview. (a) The encoder reads the context; the receiver reads a task prompt. \texttt{LatComm} connects selected layers and adds context information after each paired receiver block. Both LLMs remain frozen. (b) For one layer pair, a context compressor produces reusable slots $Z_i$. The gates compare a summary of the receiver's prompt states with the compressed slots. The reader combines the gated slot values and adds the result to the receiver state.}
    \label{fig:latcomm-overview}
    \vspace{-.1in}
\end{figure*}

\noindent\textbf{Approach and results.}
We introduce \texttt{LatComm}, a latent communication interface that separates encoding from reading (Figure~\ref{fig:latcomm-overview}). In the encoding step, a frozen context encoder LLM reads the context without the question, and learned context compressors map its token states into a fixed number of low-dimensional slots. These slots form a \emph{compressed packet}, which is computed once, stored, and sent to the receiver for every question. Because the compressors read all token states of the encoder through cross-attention, a larger packet needs only more slot queries rather than a larger projector. In the reading step, a frozen receiver LLM reads the packet through cross-attention, and gates computed from its prompt control the contribution of each slot, so that different questions can use different parts of the same packet. Both LLMs stay frozen throughout, and only the compressors, readers, and gates are trained. The interface can therefore connect the same model, models of different sizes, or models from different families. On 12 question-answering datasets, \texttt{LatComm} improves average accuracy over single-vector baselines by up to 10.2 points. On QuALITY, it reduces the receiver's input by $57\times$ and answers all questions of an article $3\times$ faster (Figure~\ref{fig:latcomm-teaser}c, d).

\noindent\textbf{Contributions.}
Our contributions are as follows:
\begin{itemize}[leftmargin=*,nosep]
    \item We formulate context communication between LLMs as two steps, encoding without the question and reading with the question, with a packet of fixed size between them.
    \item We introduce \texttt{LatComm}, an interface between two frozen LLMs with context compressors, cross-attention readers, and per-layer, per-slot gates. It is trained in two stages, so that packets stored after the first stage remain valid.
    \item We show that \texttt{LatComm} is more accurate than single-vector baselines, answers questions three times faster than text passing, and connects six LLM pairs of different sizes and families.
    \item We release a PyTorch implementation with training and evaluation configurations.\footnote{\url{https://github.com/huawei-lin/LatComm}}
\end{itemize}

\section{Related Work}
\label{sec:related-work}
\noindent\textbf{Context compression.}
Context compression reduces the input that an LLM reads. LLMLingua-2 keeps a subset of the tokens \citep{pan2024llmlingua2}, so its output is still text. Gist tokens, AutoCompressor, and ICAE train an LLM to encode a prompt or a context into a few vectors that the same LLM reads \citep{mu2023gist,chevalier2023autocompressor,ge2024icae}. xRAG maps one retrieval embedding per document into a frozen LLM \citep{cheng2024xrag}. CEPE and Flamingo feed a separately encoded input to a frozen language model through inserted cross-attention layers \citep{yen2024cepe,alayrac2022flamingo}. CEPE trains a small encoder for its decoder and keeps one vector per context token, and the gates of Flamingo are learned constants. \texttt{LatComm} compresses the hidden states of one LLM into a fixed number of slots, which a different LLM reads through gates that change with its prompt.

\noindent\textbf{Latent communication.}
Latent communication passes internal representations between models in place of text, such as activations, hidden states, translated vectors, or selected key--value (KV) pairs \citep{ramesh2025activations,DBLP:conf/acl/DuWBCZCZCY26,DBLP:journals/corr/abs-2511-03945,shi2025kvcomm}. DroidSpeak reuses KV caches across LLMs with the same architecture \citep{liu2024droidspeak}, and LatentMAS passes the KV caches of one agent to the next without training \citep{zou2025latentmas}. C2C trains cache fusers between different frozen LLMs \citep{fu2025c2c}. It transfers per-token states of an input that the two models both read, so the transfer grows with the input length. \texttt{LatComm} instead sends a packet with several slots per layer that is built without the question, and each receiver token reads it through gates computed from the prompt.

\begin{figure*}[!t]
    \centering
    \vspace{-.1in}
    \includegraphics[width=\textwidth]{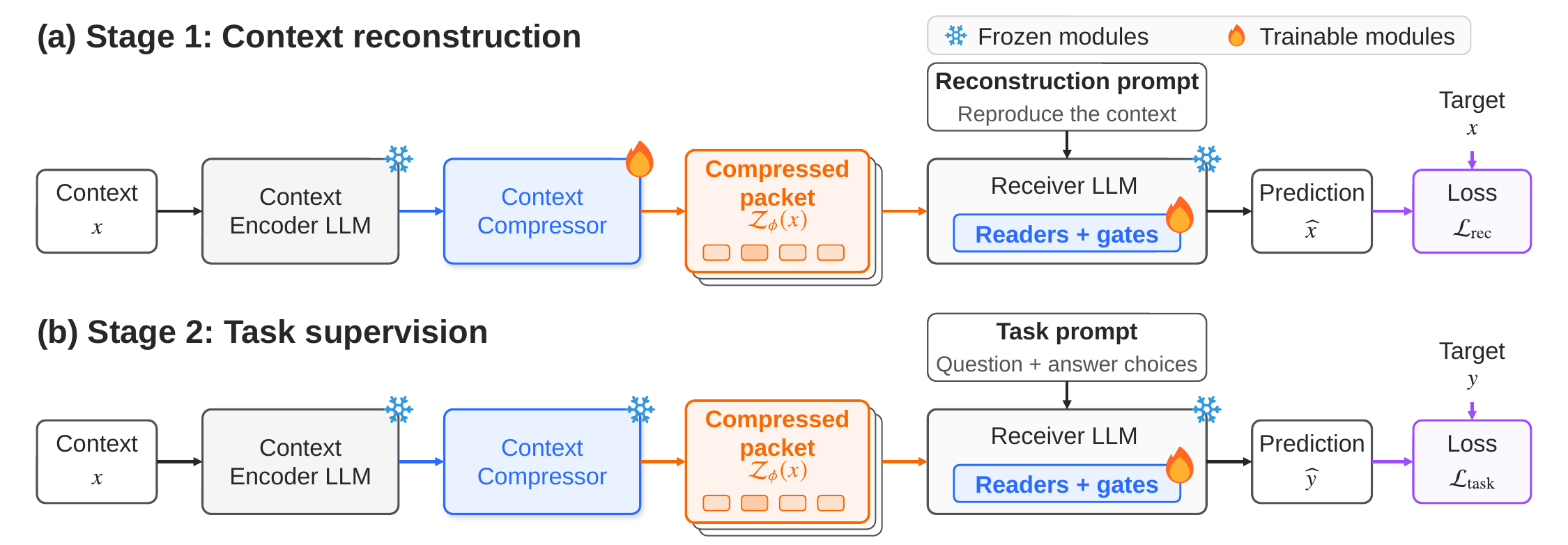}
    \vspace{-.3in}
    \caption{Two-stage training of \texttt{LatComm}. (a) In Stage 1, the receiver gets an instruction to reproduce the context $x$ and can read it only from the packet $\mathcal Z_\phi(x)$. The reconstruction loss $\mathcal L_{\mathrm{rec}}$ trains the compressors, readers, and gates. (b) In Stage 2, the receiver gets the question with its answer choices, and the task loss $\mathcal L_{\mathrm{task}}$ on the answer $y$ trains only the readers and gates. The compressors are fixed, so the packet of each training context is computed once and stored. Both LLMs stay frozen in both stages.}
    \label{fig:latcomm-training}
    \vspace{-.1in}
\end{figure*}

\section{\texttt{LatComm}: Latent Communication}
\label{sec:latcomm}

Consider a context $x$, such as a document, and a prompt $r$ that contains a question about $x$, the task instructions, and any answer choices. With text passing, the \emph{receiver LLM} reads $x$ followed by $r$ and generates the answer $y$, so it processes $x$ again for every prompt (Figure~\ref{fig:latcomm-teaser}a). With \texttt{LatComm}, the receiver reads only $r$ (Figure~\ref{fig:latcomm-teaser}b). A second LLM, the \emph{context encoder LLM}, reads $x$ once and sends a \emph{compressed packet} $\mathcal Z(x)$, a fixed number of learned vectors, and the receiver generates $y$ from $r$ and $\mathcal Z(x)$. The packet is the receiver's only source of context information, and both LLMs stay frozen.

The packet passes between selected layers of the two models (Figure~\ref{fig:latcomm-overview}a). At each of these layer pairs, a compressor turns the encoder's token states, the hidden states that the layer computes for each token, into a few slots. A reader with gates then adds the information that the current prompt needs to the receiver's states (Figure~\ref{fig:latcomm-overview}b). Only the compressors, readers, and gates are trained, in two stages.

\subsection{Encoding a Compressed Packet}
\label{sec:latcomm-compression}

We use $P$ connections, called \emph{layer pairs}. Pair $i$ connects encoder layer $a_i$ to receiver layer $b_i$. Its context compressor maps the encoder's token states $H^E_{a_i}(x)\in\mathbb R^{n_E\times d_E}$ into $S_i$ slots of dimension $d$, where $n_E$ is the number of context tokens and $d_E$ is the encoder's hidden dimension. With learned slot queries $U_i\in\mathbb R^{S_i\times d}$, compression is
\begin{equation}
    \begin{aligned}
        X_i &= H^E_{a_i}(x)W_i^E, \\
        C_i &= \operatorname{MHA}_i(U_i,X_i,X_i), \\
        B_i &= \operatorname{LN}^{\mathrm{attn}}_i(U_i+C_i), \\
        Z_i(x) &= \operatorname{LN}^{\mathrm{ffn}}_i
            \bigl(B_i+\operatorname{FFN}_i(B_i)\bigr).
    \end{aligned}
    \label{eq:packet-encoding}
\end{equation}
The projection $W_i^E$ maps context states to width $d$. Multi-head cross-attention ($\operatorname{MHA}$) lets the slot queries collect information from these states through parallel weighted reads. Layer normalization ($\operatorname{LN}$) normalizes features, and a feed-forward network ($\operatorname{FFN}$) transforms the result. Each row of $Z_i(x)\in\mathbb R^{S_i\times d}$ is one \emph{slot}. Each pair has its own compressor, and the packet $\mathcal Z(x)=(Z_1(x),\ldots,Z_P(x))$ collects the slots of all pairs.

The compressors read hidden states rather than key--value (KV) caches. A hidden state is one vector per token, whereas a KV cache holds keys and values for every attention head and is tied to the token positions of its input, so methods that transfer caches align the tokens of the two models \citep{fu2025c2c}. Slots pooled from hidden states have no token positions, and the receiver can read them without reading the context.

The encoder and the compressors see only the context, without the question or the answer choices, so the same context always produces the same packet, and the receiver can reuse it for different prompts. The packet stores $d\sum_i S_i$ numbers, which depends on the number of slots and their width but not on the number of context tokens. Because the slots come from learned queries, a larger packet needs only more queries, and the size of the compressor stays the same. The compressors are fixed after the first training stage, so each context is encoded once and its packet can be stored.

\subsection{Reading with Per-Layer, Per-Slot Gates}
\label{sec:latcomm-reading}

For each prompt, the receiver runs its usual layers, and after layer $b_i$ a trainable \emph{reader} gives it access to the slots $Z_i$. Let $H_i\in\mathbb R^{n_T\times d_T}$ be the receiver states after layer $b_i$, where $n_T$ is the number of receiver tokens and $d_T$ is the receiver's hidden dimension. The reader uses cross-attention, so each token assigns its own weights to the slots. These weights sum to one over the slots, so attention alone can shift weight between slots but cannot reduce the total contribution of the packet. A \emph{gate} $g_{i,s}\in(0,1)$ therefore sets the strength of slot $s$ at pair $i$. A learned pooling module, shared across pairs, summarizes the receiver states at the prompt positions into a vector $p_i$, and the gate compares this summary with the slot $z_{i,s}=Z_i(x)[s,:]$:
\begin{equation}
    g_{i,s}=\sigma\!\left(
        \frac{\langle p_iW_g^Q,\,z_{i,s}W_g^K\rangle}{\sqrt d}
        +\beta_{i,s}\right).
    \label{eq:slot-gate}
\end{equation}
Here $\sigma$ is the sigmoid, the projections $W_g^Q$ and $W_g^K$ are shared across pairs, and every layer--slot pair has its own bias $\beta_{i,s}$. At later layers, the prompt states already contain context information from earlier readers, so $p_i$ and the gates depend on both $x$ and $r$.

The reader projects the normalized receiver states into queries $Q_i$ and the normalized slots into keys $K_i$ and values $V_i$. It then scales the values with the gates and adds the result to the receiver states. For one attention head,
\begin{equation}
    \begin{aligned}
        A_i &= \operatorname{softmax}\bigl(Q_iK_i^\top/\sqrt d\bigr), \\
        H'_i &= H_i+A_iG_iV_iW_i^O,
    \end{aligned}
    \label{eq:reader-injection}
\end{equation}
where $G_i=\operatorname{diag}(g_{i,1},\ldots,g_{i,S_i})$. The gates scale the values and leave the attention weights unchanged, and $H'_i$ continues through the rest of the receiver. We call this \emph{dynamic per-layer, per-slot value gating}: every layer pair and slot has its own gate, and its value changes with the prompt. A gate is computed when the receiver processes the prompt, is shared by all receiver tokens, and is reused while the answer is generated. The pooling module, the multi-head form of the reader, and the initialization are specified in Appendix~\ref{sec:implementation-details}.

\subsection{Training}
\label{sec:latcomm-training}

\texttt{LatComm} is trained in two stages (Figure~\ref{fig:latcomm-training}). Stage 1, context reconstruction, teaches the interface to carry context information, and Stage 2, task supervision, teaches the receiver side to use it for a task. Let $\theta=(\phi,\psi,\eta)$ be the parameters of the compressors, readers, and gates. Both stages minimize the cross-entropy of the response tokens under teacher forcing. For a minibatch $\mathcal B$, let example $j$ have the prompt $r_j$, the context $x_j$, and the response $y_j$ with supervised positions $\mathcal R_j$, and let $\ell_{j,t}=-\log P_\theta(y_{j,t}\mid r_j,\mathcal Z_{\phi}(x_j),y_{j,<t})$ be the loss of response token $t$. The minibatch loss is
\begin{equation}
    \mathcal L(\theta;\mathcal B)=
        \frac{\sum_{j\in\mathcal B}\sum_{t\in\mathcal R_j}\ell_{j,t}}
        {\sum_{j\in\mathcal B}|\mathcal R_j|}.
    \label{eq:training-loss}
\end{equation}

\noindent\textbf{Context reconstruction.}
Stage 1 trains all of $\theta$. The receiver gets an instruction to reproduce the context, and the target is the context itself, tokenized by the receiver and followed by an end-of-sequence token. The receiver does not see the context in its prompt, so it can reproduce the context only from the packet. Predicting every context token trains the packet to keep information that different questions may need.

\noindent\textbf{Task supervision.}
Stage 2 starts from the Stage 1 weights, keeps the compressors $\phi$ fixed, and trains only the readers $\psi$ and the gates $\eta$ on the task data. For multiple-choice questions, the target is the correct option letter. Because the compressors are fixed, the packet of each training context is computed once and stored, and the context encoder LLM is not run again in this stage. For the same reason, packets that were stored before task training remain valid after it.

\begin{table}[!t]
\centering
\begingroup
\footnotesize
\setlength{\tabcolsep}{1pt}
\renewcommand{\arraystretch}{1.15}
\definecolor{lcGreen}{RGB}{28,140,51}
\definecolor{lcRed}{RGB}{196,40,40}
\newcommand{\lcYes}{\textcolor{lcGreen}{\ding{51}}}
\newcommand{\lcNo}{\textcolor{lcRed}{\ding{55}}}
\newcommand{\lcHead}[1]{{\renewcommand{\arraystretch}{0.9}\begin{tabular}[c]{@{}c@{}}#1\end{tabular}}}
\begin{tabular*}{\linewidth}{@{\extracolsep{\fill}}lcccc@{}}
\toprule
\midrule
Method & \lcHead{Compute per\\question} & \lcHead{One-time\\compute} & \lcHead{Stored\\size} & \lcHead{Cross-\\LLM} \\
\midrule
Text passing & $L_T(n_E{+}n_T)^2$ & -- & $n_E$ & \lcYes \\
KV cache & $L_T\,n_T(n_E{+}n_T)$ & $L_T\,n_E^2$ & $L_T\,n_E\,d_T$ & \lcNo \\
\textbf{\texttt{LatComm}} & $L_T\,n_T^2+P\,n_T S$ & $L_E\,n_E^2$ & $P\,S\,d$ & \lcYes \\
\midrule
\bottomrule
\end{tabular*}
\endgroup
\vspace{-.1in}
\caption{Orders of growth for three ways to reuse a context of $n_E$ tokens for questions with $n_T$-token prompts. $L_E$ and $L_T$ are the numbers of encoder and receiver layers. \texttt{LatComm} uses $P$ layer pairs with $S$ slots of width $d$. The compute columns treat hidden sizes as constants and assume full self-attention in every layer.}
\label{tab:complexity}
\vspace{-.2in}
\end{table}

\begin{table*}[!t]
\centering
\begingroup
\small
\definecolor{lcOrangeRow}{HTML}{FEF4ED}
\setlength{\tabcolsep}{2pt}
\renewcommand{\arraystretch}{1.15}
\newcommand{\lcMid}[1]{\raisebox{\dimexpr(\arraystretch\ht\strutbox+\arraystretch\dp\strutbox+\aboverulesep+\belowrulesep+\cmidrulewidth)/2\relax}[0pt][0pt]{#1}}
\resizebox{\textwidth}{!}{%
\begin{tabular}{@{}l>{\hspace{3pt}}lccccccccccccc>{\hspace{5pt}}cccc@{}}
\toprule
\midrule
 & & \multicolumn{13}{c}{{\scriptsize Task datasets (\%)}} & \multicolumn{4}{c}{{\scriptsize Held-out datasets (\%)}} \\
\cmidrule(lr){3-15}\cmidrule(l){16-19}
\lcMid{\scriptsize Method} & \lcMid{\scriptsize Encoder} & {\scriptsize CLOTH} & {\scriptsize Cosmos} & {\scriptsize HSwag} & {\scriptsize LogiQA} & {\scriptsize MCScr.} & {\scriptsize MuTual} & {\scriptsize OBQA} & {\scriptsize QuAL.} & {\scriptsize ReClor} & {\scriptsize SciQ} & {\scriptsize SIQA} & {\scriptsize SWAG} & {\scriptsize Avg} & {\scriptsize RACE} & {\scriptsize DREAM} & {\scriptsize QuAIL} & {\scriptsize Avg} \\
\midrule
\multicolumn{19}{@{}l}{\textit{Full text}} \\
\multicolumn{2}{@{}l}{Raw-text} & 45.3 & 40.8 & 37.0 & 31.0 & 82.3 & 48.0 & 68.0 & 44.9 & 38.6 & 97.5 & 48.2 & 35.6 & 51.43 & 66.1 & 70.9 & 58.2 & 65.06 \\
\midrule
\multicolumn{19}{@{}l}{\textit{No context}} \\
\multicolumn{2}{@{}l}{Query-only} & 25.0 & 44.8 & 28.9 & 26.0 & 62.6 & 32.3 & 53.6 & 32.2 & 27.8 & 77.1 & 37.7 & 27.3 & 39.60 & 39.1 & 36.2 & 34.0 & 36.41 \\
\multicolumn{2}{@{}l}{Constant packet} & 44.2 & 63.2 & 68.9 & 37.5 & 75.6 & 71.9 & 68.8 & 39.5 & 41.8 & 88.1 & 57.6 & 65.8 & 60.25 & 45.5 & 41.2 & 43.5 & 43.37 \\
\midrule
\multicolumn{19}{@{}l}{\textit{Compressed context}} \\
xRAG & SFR-Mistral & 41.2 & 55.9 & 55.1 & 33.6 & 76.0 & 52.0 & 71.2 & 40.0 & 36.4 & 89.4 & 57.9 & 61.6 & 55.87 & 47.4 & 48.8 & 43.9 & 46.70 \\
xRAG & Qwen3.5-4B & 38.5 & 53.1 & 56.1 & 36.4 & 73.7 & 58.6 & 69.6 & 37.7 & 35.4 & 88.1 & 54.8 & 61.2 & 55.27 & 44.8 & 44.8 & 44.6 & 44.71 \\
Embedding & F2LLM-v2 & 45.1 & 58.6 & 68.3 & 35.2 & 78.7 & 72.3 & 74.4 & 41.1 & 39.4 & 87.1 & 61.9 & 68.6 & 60.90 & 44.2 & 44.9 & 36.1 & 41.72 \\
Embedding & BGE-Gemma2 & 46.2 & 58.4 & 70.5 & 36.1 & 78.6 & 71.7 & 73.2 & \textbf{43.7} & 40.4 & 89.8 & 64.9 & 70.2 & 61.98 & 46.1 & 46.6 & 41.7 & 44.83 \\
\rowcolor{lcOrangeRow}
\texttt{LatComm} & Qwen3.5-4B & \textbf{52.1} & \textbf{61.9} & \textbf{73.5} & \textbf{38.6} & \textbf{80.9} & \textbf{74.0} & \textbf{79.2} & 43.1 & \textbf{49.8} & \textbf{92.4} & \textbf{66.5} & \textbf{73.1} & \textbf{65.43} & \textbf{52.2} & \textbf{52.3} & \textbf{44.8} & \textbf{49.76} \\
\midrule
\bottomrule
\end{tabular}}
\endgroup
\vspace{-.1in}
\caption{Main results with the Qwen3.5-0.8B receiver: accuracy on each dataset. Raw-text and Query-only are the untrained receiver with and without the context in its prompt, and Constant packet trains the readers and gates with one fixed packet. Avg is Task Avg or Held-out Avg. Bold marks the best compressed-context method per column.}
\label{tab:main-comparison}
\end{table*}

\begin{table}[!t]
\centering
\begingroup
\small
\setlength{\tabcolsep}{3pt}
\renewcommand{\arraystretch}{1.15}
\begin{tabular*}{\linewidth}{@{\extracolsep{\fill}}llrrrr@{}}
\toprule
\midrule
Context encoder & Receiver & $P$ & $S$ & $d$ & Packet \\
\midrule
\multicolumn{6}{@{}l}{\textit{Same model}} \\
Qwen3.5-0.8B & Qwen3.5-0.8B & 24 & 32 & 256 & 384\,KiB \\
\midrule
\multicolumn{6}{@{}l}{\textit{Larger encoder}} \\
Qwen3.5-4B & Qwen3.5-0.8B & 24 & 32 & 256 & 384\,KiB \\
Qwen3.5-9B & Qwen3.5-0.8B & 24 & 32 & 256 & 384\,KiB \\
\midrule
\multicolumn{6}{@{}l}{\textit{Cross-family}} \\
Qwen3.5-0.8B & Gemma-3-1B-pt & 24 & 32 & 256 & 384\,KiB \\
Qwen3.5-9B & Gemma-3-1B-pt & 26 & 32 & 256 & 416\,KiB \\
Granite-4.1-8B & Llama-3.2-1B & 16 & 32 & 256 & 256\,KiB \\
\midrule
\bottomrule
\end{tabular*}
\endgroup
\vspace{-.1in}
\caption{Model pairs with the default packet. $P$ is the number of layer pairs, $S$ the number of slots per pair, and $d$ the slot width. The packet stores $P\times S\times d$ values in bfloat16, which takes $2PSd$ bytes.}
\label{tab:model-pairs}
\vspace{-.2in}
\end{table}

\subsection{Complexity}
\label{sec:latcomm-complexity}

Consider a context of $n_E$ tokens that is used for $q$ questions with prompts of $n_T$ tokens, and let $c_E$ and $c_T$ be the cost per token of the context encoder and of the receiver. Reading the context as text for every question costs $C_{\mathrm{text}}$, and encoding it once costs $C_{\mathrm{packet}}$:
\begin{equation}
    \begin{aligned}
        C_{\mathrm{text}} &= q\,(n_E+n_T)\,c_T, \\
        C_{\mathrm{packet}} &= n_E\,c_E + q\,n_T\,c_T.
    \end{aligned}
    \label{eq:reuse-cost}
\end{equation}
Encoding once therefore costs less when $q>c_E/c_T$. Table~\ref{tab:complexity} compares the growth rates with text passing and with a stored KV cache of the receiver. The receiver's compute for each question grows quadratically with the context length for text and linearly for a KV cache, because each prompt token still attends to all $n_E$ context positions. For \texttt{LatComm} it does not depend on $n_E$, because each prompt token attends to $S$ slots at each of the $P$ layer pairs. The readers and gates add $O(n_T d_T d + n_T S_i d)$ work per layer pair, which Equation~\ref{eq:reuse-cost} omits. The stored size grows with $n_E$ for the cache and is fixed for the packet: with 24 layer pairs, 32 slots per pair, $d=256$, and 16-bit values, a packet takes 384\,KiB.

\section{Experiments}
\label{sec:experiments}
The experiments address three research questions. \textbf{RQ1:} Can a context be encoded once, without the question, and reused for different questions? \textbf{RQ2:} How much useful information does a packet keep under a fixed size budget? \textbf{RQ3:} Can two different LLMs communicate through the packet, across model sizes and model families?

\subsection{Setup}
\label{sec:setup}

\noindent\textbf{Datasets.}
Stage 1 uses 2,000,000 English passages from WikiText-103~\citep{merity2016pointer} and FineWeb-Edu~\citep{penedo2024fineweb}. Stage 2 uses 12 multiple-choice datasets in which each question comes with a context: CLOTH~\citep{xie-etal-2018-large}, Cosmos QA~\citep{DBLP:conf/emnlp/HuangBBC19}, HellaSwag~\citep{zellers-etal-2019-hellaswag}, LogiQA~\citep{ijcai2020p501}, MCScript2.0~\citep{ostermann-etal-2019-mcscript2}, MuTual~\citep{cui-etal-2020-mutual}, OpenBookQA~\citep{mihaylov-etal-2018-suit}, QuALITY~\citep{pang-etal-2022-quality}, ReClor~\citep{yu2020reclor}, SciQ~\citep{welbl-etal-2017-crowdsourcing}, SocialIQA~\citep{sap-etal-2019-social}, and SWAG~\citep{zellers-etal-2018-swag}. We also test on three held-out datasets not used in training: RACE~\citep{lai-etal-2017-race}, DREAM~\citep{sun-etal-2019-dream}, and QuAIL~\citep{rogers2020getting} (sizes and splits in Appendix~\ref{sec:data-details}). The prediction is the valid option letter with the highest logit at the first response position. We report the accuracy on each dataset and two unweighted means: Task Avg over the 12 task datasets and Held-out Avg over the held-out datasets.

\noindent\textbf{Model pairs.}
Table~\ref{tab:model-pairs} lists the six model pairs. The main receiver is Qwen3.5-0.8B~\citep{qwen2026qwen35}, with the same model or a larger Qwen3.5 model (4B or 9B) as the context encoder. Two more receivers cross model families: Gemma-3-1B-pt~\citep{gemma2025card} with Qwen3.5 encoders, and Llama-3.2-1B~\citep{meta2024llama32} with the Granite-4.1-8B base model~\citep{ibm2026granite41} as the encoder. Each reader is paired with the encoder layer at the same relative depth, and the default packet has 32 slots per pair of width $d=256$.

\begin{table*}[!t]
\centering
\begingroup
\small
\definecolor{lcOrangeRow}{HTML}{FEF4ED}
\setlength{\tabcolsep}{2pt}
\renewcommand{\arraystretch}{1.15}
\newcommand{\lcMid}[1]{\raisebox{\dimexpr(\arraystretch\ht\strutbox+\arraystretch\dp\strutbox+\aboverulesep+\belowrulesep+\cmidrulewidth)/2\relax}[0pt][0pt]{#1}}
\resizebox{\textwidth}{!}{%
\begin{tabular}{@{}l>{\hspace{3pt}}lccccccccccccc>{\hspace{5pt}}cccc@{}}
\toprule
\midrule
 & & \multicolumn{13}{c}{{\scriptsize Task datasets (\%)}} & \multicolumn{4}{c}{{\scriptsize Held-out datasets (\%)}} \\
\cmidrule(lr){3-15}\cmidrule(l){16-19}
\lcMid{\scriptsize Method} & \lcMid{\scriptsize Encoder} & {\scriptsize CLOTH} & {\scriptsize Cosmos} & {\scriptsize HSwag} & {\scriptsize LogiQA} & {\scriptsize MCScr.} & {\scriptsize MuTual} & {\scriptsize OBQA} & {\scriptsize QuAL.} & {\scriptsize ReClor} & {\scriptsize SciQ} & {\scriptsize SIQA} & {\scriptsize SWAG} & {\scriptsize Avg} & {\scriptsize RACE} & {\scriptsize DREAM} & {\scriptsize QuAIL} & {\scriptsize Avg} \\
\midrule
\multicolumn{19}{@{}l}{\textit{Full text}} \\
\multicolumn{2}{@{}l}{Raw-text} & 25.0 & 25.4 & 25.0 & 22.0 & 50.9 & 23.6 & 28.0 & 25.6 & 24.4 & 24.8 & 33.5 & 24.6 & 27.73 & 25.9 & 34.7 & 25.2 & 28.63 \\
\midrule
\multicolumn{19}{@{}l}{\textit{No context}} \\
\multicolumn{2}{@{}l}{Query-only} & 25.4 & 25.5 & 25.7 & 25.8 & 48.9 & 23.8 & 26.4 & 25.9 & 25.8 & 23.5 & 33.7 & 25.2 & 27.98 & 26.4 & 34.6 & 25.2 & 28.74 \\
\multicolumn{2}{@{}l}{Constant packet} & 38.2 & 59.3 & 67.9 & 28.4 & 71.8 & 59.9 & 56.8 & 35.5 & 28.2 & 82.0 & 54.8 & 65.7 & 54.04 & 39.0 & 41.6 & 43.3 & 41.30 \\
\midrule
\multicolumn{19}{@{}l}{\textit{Compressed context}} \\
Embedding & F2LLM-v2 & 37.4 & 57.8 & 64.0 & 29.6 & 71.8 & 59.4 & 57.4 & 33.1 & 30.4 & 80.4 & 57.1 & 64.9 & 53.62 & 38.8 & 40.5 & 37.7 & 38.98 \\
Embedding & BGE-Gemma2 & 37.8 & \textbf{60.1} & 68.4 & 30.0 & 72.1 & 58.5 & 60.8 & \textbf{36.5} & 33.6 & 82.0 & 58.8 & 66.9 & 55.46 & 41.2 & 41.9 & 38.6 & 40.59 \\
\rowcolor{lcOrangeRow}
\texttt{LatComm} & Qwen3.5-9B & \textbf{46.3} & 58.9 & \textbf{70.1} & \textbf{32.1} & \textbf{78.8} & \textbf{64.1} & \textbf{68.4} & 31.8 & \textbf{36.4} & \textbf{85.6} & \textbf{62.5} & \textbf{73.0} & \textbf{59.00} & \textbf{43.2} & \textbf{49.9} & \textbf{46.8} & \textbf{46.63} \\
\midrule
\bottomrule
\end{tabular}}
\endgroup
\vspace{-.1in}
\caption{Results with the Gemma-3-1B-pt receiver: accuracy on each dataset. Raw-text and Query-only are the untrained receiver with and without the context in its prompt, and both are near chance for this receiver. Avg is Task Avg or Held-out Avg. Bold marks the best compressed-context method per column.}
\label{tab:gemma-comparison}
\end{table*}

\noindent\textbf{Baselines.}
We group the compared methods by what the receiver gets. \emph{Full text:} \textbf{Raw-text} gives the frozen receiver the original context before the prompt. It is not trained on the task data, so we report it as a reference. \emph{No context:} \textbf{Query-only} gives the frozen receiver only the prompt, and \textbf{Constant packet} trains the same readers and gates as \texttt{LatComm} but gives every question the same packet, so it measures how much accuracy comes from task training alone. \emph{Compressed context:} \textbf{xRAG}~\citep{cheng2024xrag} maps one context vector to one soft token, where the vector comes from SFR-Embedding-Mistral~\citep{sfr2024embeddingmistral} or from the last-token state of Qwen3.5-4B, \texttt{LatComm}'s encoder. \textbf{Embedding} replaces the packet with one vector from a text embedding model, F2LLM-v2-14B~\citep{DBLP:journals/corr/abs-2603-19223} or BGE-Multilingual-Gemma2~\citep{bge2024gemma2}, which the same readers and gates read. Appendix~\ref{sec:baseline-details} describes each baseline. All trained methods share the task data, loss, batch size, epochs, and learning-rate schedule of Table~\ref{tab:hyperparameters}.

\subsection{Main Results}
\label{sec:main-results}

Table~\ref{tab:main-comparison} compares all methods with the Qwen3.5-0.8B receiver on the 12 task datasets and on the three held-out datasets. \texttt{LatComm} is the best compressed-context method on both averages. On the task datasets, it reaches 65.43 Task Avg, against 61.98 for the best baseline, and it has the highest accuracy among the compressed-context methods on 11 of the 12 datasets. On the held-out datasets, which are not used in training, it reaches 49.76 Held-out Avg, against 46.70 for the best baseline, and it has the highest accuracy on all three datasets. \texttt{LatComm} is also above the Constant packet by 5.2 points on Task Avg and by 6.4 points on Held-out Avg. The Constant packet uses the same trained readers and gates but carries no context information, so this gain shows that the packet gives the receiver useful information about the context.

Raw-text is not trained on the task data. Every trained method is above Raw-text on the task datasets and below it on the held-out datasets (41.72--49.76 against 65.06 Held-out Avg), and this includes the Constant packet. The modules trained on the 12 task datasets therefore transfer only in part to new datasets. This holds for every trained method, and \texttt{LatComm} is the most accurate.

\subsection{Cross-Family Receiver}
\label{sec:gemma-results}

Table~\ref{tab:gemma-comparison} repeats the comparison with a receiver from another model family, Gemma-3-1B-pt, and with the Qwen3.5-9B context encoder. Without training, this receiver is close to chance level with or without the context (27.73 and 27.98 Task Avg). \texttt{LatComm} reaches 59.00 Task Avg, against 55.46 for the best baseline and 54.04 for the Constant packet. On the held-out datasets, \texttt{LatComm} reaches 46.63 Held-out Avg, against 40.59 for the best baseline and 41.30 for the Constant packet. It has the highest accuracy among the compressed-context methods on 10 of the 12 task datasets and on all three held-out datasets. The interface therefore also works when the context encoder and the receiver come from different model families.

The two Embedding baselines are below the Constant packet on Held-out Avg. This difference comes from QuAIL. Every QuAIL question has a ``not enough information'' option, which is correct for 306 of the 2,720 questions. The Embedding baselines choose it for 18.5\% and 25.0\% of the questions, against 7.4\% for the Constant packet and 7.5\% for \texttt{LatComm}. On RACE and DREAM, the Embedding baselines are within 2.3 points of the Constant packet.

\subsection{Cross-LLM Communication}
\label{sec:encoder-choice}
\begin{table}[!t]
\centering
\begingroup
\footnotesize
\definecolor{lcOrangeRow}{HTML}{FEF4ED}
\edef\lcAbove{\the\aboverulesep}
\edef\lcBelow{\the\belowrulesep}
\setlength{\aboverulesep}{0pt}
\setlength{\belowrulesep}{0pt}
\newcommand{\lcT}{\rule{0pt}{\dimexpr 1.15\ht\strutbox+\lcBelow\relax}}
\newcommand{\lcB}{\rule[-\dimexpr 1.15\dp\strutbox+\lcAbove\relax]{0pt}{0pt}}
\setlength{\tabcolsep}{3pt}
\renewcommand{\arraystretch}{1.15}
\resizebox{\linewidth}{!}{%
\begin{tabular}{@{}llcc>{\columncolor{lcOrangeRow}[\tabcolsep][0pt]}c@{}}
\toprule
\addlinespace[\doublerulesep]
\midrule
\lcT \lcB Encoder & Receiver & Constant packet & Embedding & \texttt{LatComm} \\
\midrule
\multicolumn{4}{@{}l}{\lcT \textit{Same model}} & \\
\lcB Qwen3.5-0.8B & Qwen3.5-0.8B & 60.25 & 61.98 & \textbf{64.91} \\
\midrule
\multicolumn{4}{@{}l}{\lcT \textit{Larger encoder}} & \\
Qwen3.5-4B & Qwen3.5-0.8B & 60.25 & 61.98 & \textbf{65.43} \\
\lcB Qwen3.5-9B & Qwen3.5-0.8B & 60.25 & 61.98 & \textbf{65.80} \\
\midrule
\multicolumn{4}{@{}l}{\lcT \textit{Cross-family}} & \\
Qwen3.5-0.8B & Gemma-3-1B-pt & 54.04 & 55.46 & \textbf{59.21} \\
Qwen3.5-9B & Gemma-3-1B-pt & 54.04 & 55.46 & \textbf{59.00} \\
\lcB Granite-4.1-8B & Llama-3.2-1B & 60.62 & 61.42 & \textbf{64.08} \\
\midrule
\addlinespace[\doublerulesep]
\bottomrule
\end{tabular}}
\endgroup
\vspace{-.1in}
\caption{Task Avg (\%) of six model pairs and two trained references with the same receiver. Embedding uses BGE-Multilingual-Gemma2. Each receiver's Constant packet was trained with its 4B, 9B, or Granite pair.}
\label{tab:encoder-choice}
\vspace{-.1in}
\end{table}

For RQ3, Table~\ref{tab:encoder-choice} compares \texttt{LatComm} with two trained references for six pairs of LLMs. In every pair, the receiver does not read the context, so its context information comes only from the packet that the context encoder LLM sends. With the Qwen3.5-0.8B receiver, Task Avg is 64.91 when the same model is the context encoder, 65.43 with the 4B encoder, and 65.80 with the 9B encoder. The three encoders are within 0.9 points of each other, so the interface works with a small encoder, and one model can serve as both encoder and receiver. Across model families, the Gemma-3-1B-pt receiver reaches 59.00--59.21 with Qwen3.5 encoders, and the Llama-3.2-1B receiver reaches 64.08 with the Granite-4.1-8B encoder. In all six pairs, \texttt{LatComm} is 3.5 to 5.6 points above the Constant packet and 2.7 to 3.8 points above the Embedding baseline, both trained with the same receiver. The packet thus passes context information between LLMs of different sizes and families.

The receiver's answers depend on the packet it receives. With the Qwen3.5-4B encoder and the 0.8B receiver, the same readers and gates reach 65.4 Task Avg with the packet of the question's own context, 45.6 with the packet of another context, and 39.6 without a packet (Table~\ref{tab:packet-controls}).

\subsection{Reuse Cost}
\label{sec:reuse-cost}
\begin{table}[!t]
\centering
\begingroup
\small
\definecolor{lcOrangeRow}{HTML}{FEF4ED}
\edef\lcAbove{\the\aboverulesep}
\edef\lcBelow{\the\belowrulesep}
\setlength{\aboverulesep}{0pt}
\setlength{\belowrulesep}{0pt}
\newcommand{\lcT}{\rule{0pt}{\dimexpr 1.15\ht\strutbox+\lcBelow\relax}}
\newcommand{\lcB}{\rule[-\dimexpr 1.15\dp\strutbox+\lcAbove\relax]{0pt}{0pt}}
\setlength{\tabcolsep}{4pt}
\renewcommand{\arraystretch}{1.15}
\begin{tabular*}{\linewidth}{@{\extracolsep{\fill}}lc>{\columncolor{lcOrangeRow}[\tabcolsep][0pt]}c@{}}
\toprule
\addlinespace[\doublerulesep]
\midrule
\lcT \lcB Quantity & Raw-text & \texttt{LatComm} \\
\midrule
\multicolumn{2}{@{}l}{\lcT \textit{Per question}} & \\
Receiver input tokens & 5,915 & 103 \\
Receiver time (ms) & 511 & 131 \\
\lcB Receiver peak memory (GiB) & 0.87 & 0.06 \\
\midrule
\multicolumn{2}{@{}l}{\lcT \textit{Per context}} & \\
Encoding time (s) & -- & 0.71 \\
\lcB Total time, 18.1 questions (s) & 9.26 & 3.09 \\
\midrule
\addlinespace[\doublerulesep]
\bottomrule
\end{tabular*}
\endgroup
\vspace{-.1in}
\caption{Measured cost on the QuALITY test set (115 articles, 2,086 questions) with the Qwen3.5-4B encoder and the Qwen3.5-0.8B receiver. Times are measured on one H100 GPU at batch size 1. Peak memory is the GPU memory that one call adds to the loaded models.}
\label{tab:reuse-cost}
\vspace{-.1in}
\end{table}

\begin{figure}[!t]
    \centering
    \includegraphics[width=\linewidth]{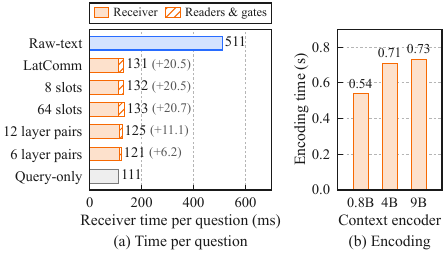}
    \caption{Reuse cost on QuALITY. (a) Receiver time per question for Raw-text, five \texttt{LatComm} settings, and Query-only. The hatched part of a bar is the time added by the readers and gates. (b) Encoding time per context for three context encoders.}
    \label{fig:reuse-cost}
\end{figure}

For RQ1, one packet serves every question about its context. In the test set, one context has 18.1 questions on average in QuALITY, 14.2 in CLOTH, and 5.7 in MCScript2.0. Table~\ref{tab:reuse-cost} reports the measured cost on QuALITY. Encoding an article once takes 0.71\,s. The receiver then answers a question in 131\,ms with the packet, against 511\,ms from the full text (Figure~\ref{fig:latcomm-teaser}d). A Query-only pass takes 111\,ms, so the readers and gates add only 20\,ms per question. For one article with 18.1 questions, the packet path takes 3.09\,s in total and the text path takes 9.26\,s, so the packet path is 3.0 times faster (Figure~\ref{fig:latcomm-teaser}c). Encoding once costs less from the second question.

The gain grows with the context length (Figure~\ref{fig:reuse-cost-length}). With text passing, the time per question grows from 300\,ms for the fifth of the questions with the shortest articles to 651\,ms for the fifth with the longest. With a packet it stays near 131\,ms, so the receiver is 2.3 to 4.9 times faster. Figure~\ref{fig:reuse-cost}a compares the settings of \texttt{LatComm} with Raw-text and Query-only. Every setting takes between 121 and 133\,ms per question. The slot count does not change the time, and with 6 layer pairs the readers and gates add 6.2\,ms instead of 20.5\,ms. Encoding takes 0.54 to 0.73\,s with the three encoders (Figure~\ref{fig:reuse-cost}b).

For RQ2, the default packet takes 384\,KiB per context. Task Avg stays between 65.0 and 65.8 when the slot count changes from 8 to 64, so a 96\,KiB packet with 8 slots per pair keeps the accuracy of the default packet. A slot width of 512 raises Task Avg to 66.88. Every packet configuration is above the best single-vector baseline (61.98). Appendix~\ref{sec:ablation} reports all configurations.

\section{Conclusion}
\label{sec:conclusion}
We presented \texttt{LatComm}, a latent communication interface between frozen LLMs. A context encoder LLM compresses a context once into a fixed-size packet, and a receiver LLM, which can differ from the encoder in size and family, reads it for every question through cross-attention with gates computed from its prompt. \texttt{LatComm} improves average accuracy over single-vector baselines by up to 10.2 points, and on QuALITY it answers the questions of an article $3\times$ faster than text passing. Future work will study packets as messages in multi-agent systems and as entries in agent memory.

\clearpage
\section*{Limitations}
Our experiments cover multiple-choice question answering in English, where the answer is a single option letter. We do not evaluate free-form generation, and we do not test the multi-agent and agent-memory uses that motivate reusable packets. The receivers have about one billion parameters and the encoders have at most nine billion, so the behavior of the interface with larger receivers is not known. In addition, only Stage 2 of the default pair is repeated with three seeds, which gives a standard deviation of 0.23 points in Task Avg. Every other configuration uses one seed, so small differences between the configurations of the ablation study should be read with care.

\texttt{LatComm} is trained for one pair of models, so a new context encoder or receiver needs its own compressors, readers, and gates, and its own packets. The interface also reads and writes hidden states, and therefore it does not apply to models that are available only through a text interface. Finally, because the packet has a fixed size, a longer context is compressed more strongly. Stage 1 uses passages of at most 1,024 tokens, and the longest contexts in our evaluation are the QuALITY articles. We do not study how the packet size should grow with the context length.

\bibliography{references}

\appendix
\section{Implementation Details}
\label{sec:implementation-details}

\noindent\textbf{Layer pairing.}
Equations use row vectors and omit the batch dimension. Let $n_E,n_T$ be the context encoder and receiver token counts, and $d_E,d_T$ their hidden dimensions. We choose $P$ unique layer pairs $(a_i,b_i)$, with blocks selected uniformly across each model's depth and paired in order. When $P>1$, the schedule includes the final blocks. Each pair has its own compressor and reader, and may have a different slot count $S_i$. In the six model pairs of Table~\ref{tab:model-pairs}, every receiver layer has a reader, with one exception: the Qwen3.5-0.8B encoder has 24 layers and the Gemma-3-1B-pt receiver has 26, so 24 receiver layers have a reader. The trainable parameters are $\theta=(\phi,\psi,\eta)$ for context compressors, readers, and gates, respectively. Both backbone parameter sets are fixed.

\noindent\textbf{Context compressor.}
Equation~\ref{eq:packet-encoding} uses $W_i^E\in\mathbb R^{d_E\times d}$ and masks encoder padding positions in cross-attention. The learned-query residual is retained. The feed-forward network expands to $2d$, applies SiLU, and projects back to $d$:
\begin{equation}
    \operatorname{FFN}_i(B)=
        \operatorname{SiLU}(BW_{i,1}+b_{i,1})W_{i,2}+b_{i,2}.
    \label{eq:compressor-details}
\end{equation}
Here $W_{i,1}\in\mathbb R^{d\times2d}$, $W_{i,2}\in\mathbb R^{2d\times d}$, $b_{i,1}\in\mathbb R^{1\times2d}$, and $b_{i,2}\in\mathbb R^{1\times d}$. Compressors and readers use the same attention-head count $M$, which must divide $d$. The implementation sets $M=8$.

\noindent\textbf{Pooling and gates.}
Let $\mathcal I(r)$ be the unmasked prompt positions, and $h_{i,t}$ the receiver block output before the current reader. A shared attention pool summarizes these states:
\begin{align}
    \bar h_{i,t} &= \operatorname{LN}_g(h_{i,t}), \\
    e_{i,t} &= \tanh(\bar h_{i,t}W_p+b_p)w_p, \\
    \alpha_{i,t} &= \frac{\exp(e_{i,t})}
        {\sum_{v\in\mathcal I(r)}\exp(e_{i,v})}, \\
    p_i(x,r) &= \sum_{t\in\mathcal I(r)}\alpha_{i,t}\bar h_{i,t}.
    \label{eq:gate-pooling}
\end{align}
Here $W_p\in\mathbb R^{d_T\times d_p}$, $b_p\in\mathbb R^{1\times d_p}$, and $w_p\in\mathbb R^{d_p\times1}$, where $d_p$ is the hidden width of the pooling network. LayerNorm and pooling weights are trainable and shared across all pairs. Response and padding positions have zero pooling weight. The summary pools normalized contextual states rather than input embeddings.
The gate projections are
\begin{align}
    u_i(x,r) &= p_i(x,r)W_g^Q, \\
    k_{i,s}(x) &= z_{i,s}(x)W_g^K.
    \label{eq:gate-projections}
\end{align}
Here $p_i(x,r)\in\mathbb R^{1\times d_T}$, $z_{i,s}(x)=Z_i(x)[s,:]\in\mathbb R^{1\times d}$, $W_g^Q\in\mathbb R^{d_T\times d}$, and $W_g^K\in\mathbb R^{d\times d}$. These bias-free projections are shared across all pairs and slots.
These projections give the prompt--slot comparison in Equation~\ref{eq:slot-gate}. Gate keys $k_{i,s}$ use a separate projection from reader attention keys $K_i$.
Independent biases $\beta_{i,s}$ are initialized to $2$. Including pooling, the gate module has $d_Td+d^2+\sum_iS_i+d_Td_p+2d_p+2d_T$ parameters. The implementation sets $d_p=256$ and computes pooling and gate logits in bfloat16, without an additional LLM forward pass. Inference caches gates separately for each paired layer during prompt prefill; training retains the contextual gradient path, including earlier readers.

\noindent\textbf{Reader.}
In Equation~\ref{eq:reader-injection}, $W_i^Q\in\mathbb R^{d_T\times d}$ and $W_i^K,W_i^V\in\mathbb R^{d\times d}$. All three projections are bias-free; receiver and latent LayerNorms are separate. For head $m$, $Q_i^{(m)}\in\mathbb R^{n_T\times d_h}$ and $K_i^{(m)},V_i^{(m)}\in\mathbb R^{S_i\times d_h}$. Attention has shape $A_i^{(m)}\in\mathbb R^{n_T\times S_i}$, and $G_i\in\mathbb R^{S_i\times S_i}$. Thus the gated read in Equation~\ref{eq:reader-injection} is well-defined and produces $R_i^{(m)}\in\mathbb R^{n_T\times d_h}$. Writing $a_{i,t,s}^{(m)}=A_i^{(m)}[t,s]$ and $v_{i,s}^{(m)}=V_i^{(m)}[s,:]$, the output for token $t$ is
\begin{equation}
    r_{i,t}^{(m)}=\sum_{s=1}^{S_i}
        a_{i,t,s}^{(m)}g_{i,s}v_{i,s}^{(m)}.
    \label{eq:reader-attention}
\end{equation}
Attention coefficients sum to one across slots; their gated coefficients are not renormalized. The same $G_i$ is used for all tokens and heads. It is notation for row-wise scaling, implemented by broadcasting rather than constructing a dense matrix. The output projection $W_i^O\in\mathbb R^{d\times d_T}$ is bias-free and initialized to zero, so the initial residual update is zero. Attention dropout is zero.

\noindent\textbf{Supervision and payload.}
Reconstruction targets include the end-of-sequence token. Equation~\ref{eq:training-loss} averages over supervised response tokens, rather than giving equal weight to responses of different lengths. The context encoder LLM is evaluated without gradients. In Stage 1, gradients through the frozen receiver update the compressors, readers, and gates. In Stage 2, they update only the readers and gates. If a stored scalar occupies $w$ bytes, the packet's payload is $wd\sum_iS_i$ bytes, excluding metadata. Gate and reader parameters are stored at the receiver rather than transmitted in the packet.

\noindent\textbf{Training protocol implementation.}
Stage 2 loads the complete Stage 1 interface, including the prompt-pooling and gate parameters, from the final Stage 1 checkpoint with strict key matching, and starts a new optimizer and learning-rate schedule. The model pair, layer pairing, slot count, and latent width are identical in both stages, so every interface tensor transfers without reshaping or reinitialization. The compressor parameters are then frozen. Packets for the training contexts are computed once with the frozen encoder and compressors; identical contexts share one stored packet. The packets are stored in bfloat16. Stage 2 training reads the packets and does not load the context encoder LLM.

\begin{table}[!t]
\centering
\begingroup
\small
\setlength{\tabcolsep}{3pt}
\renewcommand{\arraystretch}{1.2}
\newcommand{\lcHead}[1]{{\renewcommand{\arraystretch}{0.9}\begin{tabular}[c]{@{}c@{}}#1\end{tabular}}}
\begin{tabularx}{\linewidth}{@{}Xrrrr@{}}
\toprule
\midrule
Dataset & \lcHead{Train\\examples} & \lcHead{Train context\\tokens} & \lcHead{Avg.\\context} & Eval \\
\midrule
\multicolumn{5}{@{}l}{\textit{Stage 1: context reconstruction}} \\
WikiText-103 & 446,282 & 100,278,346 & 224.7 & 1,057 \\
FineWeb-Edu & 1,553,718 & 1,033,473,862 & 665.2 & 13,926 \\
\textbf{Stage 1 total} & \textbf{2,000,000} & \textbf{1,133,752,208} & \textbf{566.9} & \textbf{14,983} \\
\midrule
\multicolumn{5}{@{}l}{\textit{Stage 2: task supervision}} \\
CLOTH & 76,660 & 34,320,807 & 447.7 & 11,516 \\
Cosmos QA & 25,067 & 1,861,200 & 74.2 & 2,985 \\
HellaSwag & 31,308 & 1,472,101 & 47.0 & 10,042 \\
LogiQA & 7,202 & 578,041 & 80.3 & 651 \\
MCScript2.0 & 14,017 & 2,372,120 & 169.2 & 3,610 \\
MuTual & 6,901 & 685,106 & 99.3 & 886 \\
OpenBookQA & 3,205 & 33,160 & 10.3 & 500 \\
QuALITY & 2,393 & 14,215,251 & 5,940.3 & 2,086 \\
ReClor & 4,422 & 351,546 & 79.5 & 500 \\
SciQ & 10,210 & 992,369 & 97.2 & 884 \\
SocialIQA & 33,232 & 554,725 & 16.7 & 2,224 \\
SWAG & 67,824 & 1,118,265 & 16.5 & 20,006 \\
\textbf{Stage 2 total} & \textbf{282,441} & \textbf{58,554,691} & \textbf{207.3} & \textbf{55,890} \\
\midrule
\multicolumn{5}{@{}l}{\textit{Held-out: test only}} \\
RACE & -- & -- & 343.6 & 4,934 \\
DREAM & -- & -- & 156.4 & 2,041 \\
QuAIL & -- & -- & 419.6 & 2,720 \\
\midrule
\bottomrule
\end{tabularx}
\endgroup
\caption{Data for both training stages and the held-out datasets. Train excludes the development holdout. Tokens count untruncated context text once per training example with the pinned Qwen3.5-0.8B tokenizer, excluding instructions, choices, targets, and special tokens. Reconstruction supervision contains the same text plus one EOS per passage (1,135,752,208 target tokens). Avg. context is the mean training context length in tokens, rounded to one decimal; total-row means are weighted by example count. For the held-out datasets, it is the mean context length of the test questions. Eval is the reconstruction test set, the fixed public multiple-choice reporting suite, or the held-out test set.}
\label{tab:training-data}
\end{table}

\noindent\textbf{Training settings.}
Table~\ref{tab:hyperparameters} lists the training hyperparameters. Stage 1 runs one epoch over the reconstruction passages. Stage 2 then fixes the compressors, stores one packet for each training context, and runs three epochs over the task training set. The readers and gates that it trains have 16.4M parameters for the Qwen3.5-0.8B receiver. Both stages use AdamW with the default momentum coefficients. In Stage 1, the learning rate rises linearly for 157 updates and then stays constant. In Stage 2, it rises linearly for 45 updates and then follows a half-cosine decay to zero. The trained baselines use the same task data, loss, batch size, number of epochs, and learning-rate schedule. Every run uses random seed 42, and Stage 2 of the default pair is repeated with seeds 43 and 44.

\begin{table}[!t]
\centering
\begingroup
\small
\setlength{\tabcolsep}{4pt}
\renewcommand{\arraystretch}{1.15}
\begin{tabular*}{\linewidth}{@{\extracolsep{\fill}}lcc@{}}
\toprule
\midrule
Setting & Stage 1 & Stage 2 \\
\midrule
\multicolumn{3}{@{}l}{\textit{Data}} \\
Training examples & 2,000,000 & 282,441 \\
Epochs & 1 & 3 \\
Updates & 5,209 & 2,208 \\
Batch size & 384 & 384 \\
\midrule
\multicolumn{3}{@{}l}{\textit{Optimization}} \\
Trained modules & all & readers, gates \\
Optimizer & AdamW & AdamW \\
Peak learning rate & $10^{-4}$ & $10^{-4}$ \\
Warmup updates & 157 & 45 \\
Schedule after warmup & constant & half-cosine to 0 \\
Weight decay & 0 & 0 \\
Gradient-norm clipping & 1.0 & 1.0 \\
Random seed & 42 & 42 \\
\midrule
\multicolumn{3}{@{}l}{\textit{Interface}} \\
Slots per layer pair $S$ & \multicolumn{2}{c}{32} \\
Slot width $d$ & \multicolumn{2}{c}{256} \\
Attention heads $M$ & \multicolumn{2}{c}{8} \\
Pooling width $d_p$ & \multicolumn{2}{c}{256} \\
Initial gate bias $\beta_{i,s}$ & \multicolumn{2}{c}{2} \\
\midrule
\multicolumn{3}{@{}l}{\textit{Precision}} \\
Frozen LLMs & \multicolumn{2}{c}{bfloat16} \\
Interface modules & \multicolumn{2}{c}{bfloat16} \\
\midrule
\bottomrule
\end{tabular*}
\endgroup
\caption{Training hyperparameters of the default configuration. Stage 1 trains the compressors, readers, and gates, and Stage 2 trains only the readers and gates.}
\label{tab:hyperparameters}
\end{table}

\noindent\textbf{Environment.}
\texttt{LatComm} runs use Python 3.12, PyTorch 2.11 or 2.13, and Hugging Face Transformers 5.15.1. The frozen models and the interface modules run in 16-bit bfloat16, and packets are stored in the same format. All experiments run on six machines, each with eight NVIDIA H100 GPUs (80\,GB each). The trained interfaces are evaluated with Transformers. The Raw-text reference on the 12 task datasets is computed with vLLM.

\section{Data Details}
\label{sec:data-details}

\noindent\textbf{Reconstruction data.}
The Stage 1 training set contains 2,000,000 passages: 446,282 from WikiText-103 and 1,553,718 from FineWeb-Edu. They are a fixed selection of complete passages from the prepared corpus. Each passage contains 64--1,024 tokens under the Qwen3.5-0.8B tokenizer. The receiver is asked to reproduce the passage with one of 100 reconstruction instructions. Validation and test contain 15,043 and 14,983 passages. WikiText-103 keeps its official splits, and FineWeb-Edu is split by canonical URL before passage extraction. English filtering, passage deduplication, and document-level separation are applied.

\begin{table}[!t]
\centering
\begingroup
\footnotesize
\setlength{\tabcolsep}{3pt}
\renewcommand{\arraystretch}{1.15}
\newcommand{\lcMid}[1]{\raisebox{\dimexpr(\arraystretch\ht\strutbox+\arraystretch\dp\strutbox+\aboverulesep+\belowrulesep+\cmidrulewidth)/2\relax}[0pt][0pt]{#1}}
\resizebox{\linewidth}{!}{%
\begin{tabular}{@{}lccccc@{}}
\toprule
\midrule
 & & & \multicolumn{2}{c}{Reconstruction} & \\
\cmidrule(lr){4-5}
\lcMid{Setting} & \lcMid{Packet} & \lcMid{Size} & Loss & Acc. & \lcMid{Task Avg} \\
\midrule
\multicolumn{6}{@{}l}{\textit{Encoder}} \\
Qwen3.5-0.8B & $24\times32\times256$ & 384\,KiB & 1.83 & 56.3 & 64.91 \\
Qwen3.5-4B (default) & $24\times32\times256$ & 384\,KiB & 1.24 & 68.5 & 65.43 \\
Qwen3.5-9B & $24\times32\times256$ & 384\,KiB & 0.95 & 75.1 & 65.80 \\
\midrule
\multicolumn{6}{@{}l}{\textit{Slots per pair}} \\
8 & $24\times8\times256$ & 96\,KiB & 1.58 & 61.3 & 65.81 \\
32 (default) & $24\times32\times256$ & 384\,KiB & 1.24 & 68.5 & 65.43 \\
64 & $24\times64\times256$ & 768\,KiB & 1.11 & 71.5 & 65.04 \\
\midrule
\multicolumn{6}{@{}l}{\textit{Slot width}} \\
128 & $24\times32\times128$ & 192\,KiB & 1.82 & 57.1 & 65.01 \\
256 (default) & $24\times32\times256$ & 384\,KiB & 1.24 & 68.5 & 65.43 \\
512 & $24\times32\times512$ & 768\,KiB & 0.78 & 79.0 & 66.88 \\
\midrule
\multicolumn{6}{@{}l}{\textit{Layer pairs}} \\
6 & $6\times32\times256$ & 96\,KiB & 1.77 & 57.9 & 63.84 \\
12 & $12\times32\times256$ & 192\,KiB & 1.47 & 63.5 & 64.76 \\
24 (default) & $24\times32\times256$ & 384\,KiB & 1.24 & 68.5 & 65.43 \\
\midrule
\bottomrule
\end{tabular}}
\endgroup
\caption{Ablation of the context encoder and the packet settings with the Qwen3.5-0.8B receiver. Each group changes one setting of the default configuration. Packet gives $P\times S\times d$: layer pairs, slots per pair, and slot width. Size is the stored size. Reconstruction gives the loss and the token accuracy (\%) on 256 held-out passages at the end of Stage 1. Task Avg (\%) is measured after Stage 2.}
\label{tab:ablation}
\end{table}

\begin{figure*}[!t]
    \centering
    \includegraphics[width=\textwidth]{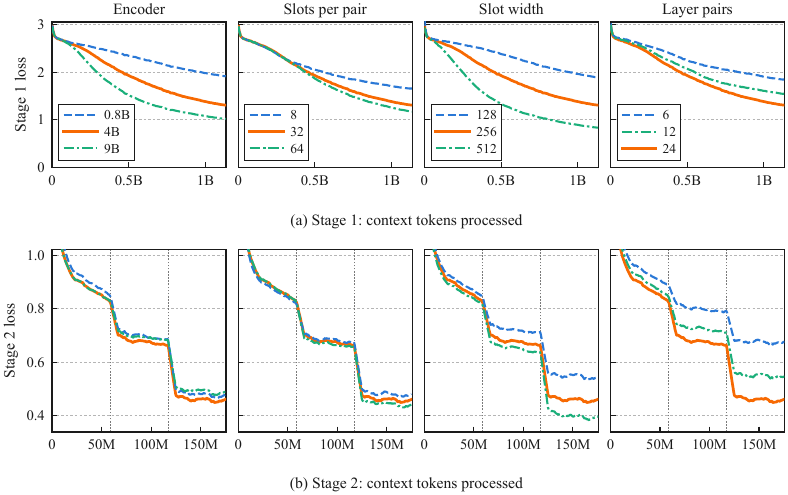}
    \caption{Training loss of (a) Stage 1 and (b) Stage 2 against the number of context tokens processed, for the configurations of Table~\ref{tab:ablation}. Each column changes one setting, and the solid orange line is the default configuration. The token count is the number of processed examples times the mean context length of the training set. The curves are running means over 50 updates in Stage 1 and 100 updates in Stage 2. Dotted lines mark the ends of the first and second epochs of Stage 2.}
    \label{fig:training-loss}
\end{figure*}

\begin{table*}[!t]
\centering
\begingroup
\small
\setlength{\tabcolsep}{2pt}
\renewcommand{\arraystretch}{1.15}
\newcommand{\lcMid}[1]{\raisebox{\dimexpr(\arraystretch\ht\strutbox+\arraystretch\dp\strutbox+\aboverulesep+\belowrulesep+\cmidrulewidth)/2\relax}[0pt][0pt]{#1}}
\resizebox{\textwidth}{!}{%
\begin{tabular}{@{}lccccccccccccc>{\hspace{5pt}}cccc@{}}
\toprule
\midrule
 & \multicolumn{13}{c}{{\scriptsize Task datasets (\%)}} & \multicolumn{4}{c}{{\scriptsize Held-out datasets (\%)}} \\
\cmidrule(lr){2-14}\cmidrule(l){15-18}
\lcMid{\scriptsize Packet given to the receiver} & {\scriptsize CLOTH} & {\scriptsize Cosmos} & {\scriptsize HSwag} & {\scriptsize LogiQA} & {\scriptsize MCScr.} & {\scriptsize MuTual} & {\scriptsize OBQA} & {\scriptsize QuAL.} & {\scriptsize ReClor} & {\scriptsize SciQ} & {\scriptsize SIQA} & {\scriptsize SWAG} & {\scriptsize Avg} & {\scriptsize RACE} & {\scriptsize DREAM} & {\scriptsize QuAIL} & {\scriptsize Avg} \\
\midrule
\multicolumn{18}{@{}l}{\textit{Same trained readers and gates}} \\
Own packet & 52.2 & 61.8 & 73.4 & 37.9 & 80.8 & 73.9 & 79.6 & 43.2 & 49.8 & 92.3 & 66.5 & 73.2 & 65.39 & 52.3 & 52.5 & 44.6 & 49.82 \\
Packet of another context & 30.7 & 48.1 & 50.3 & 28.1 & 63.4 & 43.8 & 51.2 & 31.7 & 30.6 & 77.3 & 45.4 & 46.4 & 45.59 & 37.8 & 38.5 & 32.5 & 36.28 \\
No packet & 25.0 & 44.8 & 28.9 & 26.0 & 62.6 & 32.3 & 53.6 & 32.2 & 27.8 & 77.1 & 37.7 & 27.3 & 39.60 & 39.1 & 36.2 & 34.0 & 36.41 \\
\midrule
\multicolumn{18}{@{}l}{\textit{Readers and gates trained with one fixed packet}} \\
Constant packet & 44.2 & 63.2 & 68.9 & 37.5 & 75.6 & 71.9 & 68.8 & 39.5 & 41.8 & 88.1 & 57.6 & 65.8 & 60.25 & 45.5 & 41.2 & 43.5 & 43.37 \\
\midrule
\bottomrule
\end{tabular}}
\endgroup
\caption{Packet controls for the default configuration: accuracy on each dataset. The first three rows use the same trained readers and gates and change only the packet. The packet of another context comes from a different test question. The three rows come from separate evaluation runs, so the first row differs slightly from Table~\ref{tab:main-comparison}. Columns and abbreviations follow Table~\ref{tab:main-comparison}.}
\label{tab:packet-controls}
\end{table*}

\begin{figure}[!t]
    \centering
    \includegraphics[width=\linewidth]{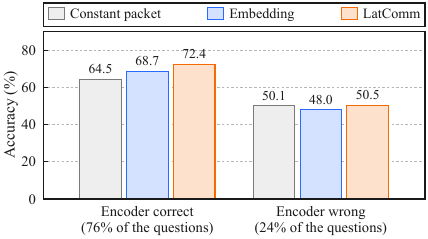}
    \caption{Accuracy on the test questions that the Qwen3.5-4B encoder answers correctly or wrongly from the full text. Embedding uses BGE-Gemma2.}
    \label{fig:gain-source}
\end{figure}

\begin{figure}[!t]
    \centering
    \includegraphics[width=\linewidth]{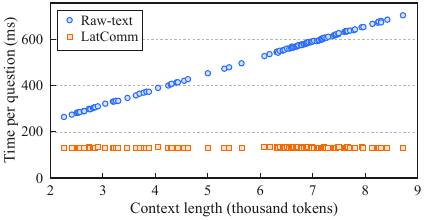}
    \caption{Receiver time per question against the context length on QuALITY. Each point is one article.}
    \label{fig:reuse-cost-length}
\end{figure}

\begin{figure*}[!t]
    \centering
    \includegraphics[width=\textwidth]{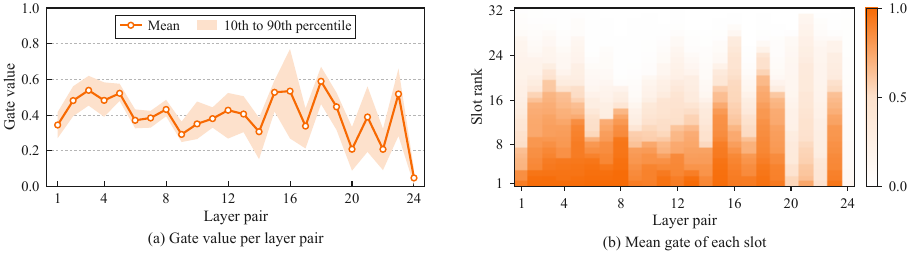}
    \caption{Gate values of the default pair on 5,386 test questions. (a) Mean over the 32 slots of each layer pair. The band covers the 10th to 90th percentile of the questions. (b) Mean gate of each slot. The slots of a layer pair are sorted from the highest mean gate to the lowest.}
    \label{fig:gate-values}
\end{figure*}

\begin{figure*}[!t]
    \centering
    \includegraphics[width=\textwidth]{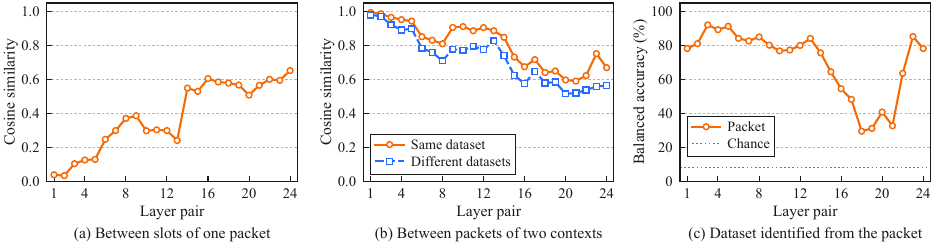}
    \caption{Packets of the default pair at each layer pair. (a) Mean cosine similarity between two slots of one packet. (b) Mean cosine similarity between the packets of two contexts from the same dataset or from different datasets. (c) Balanced accuracy of identifying the source dataset of a context from its packet.}
    \label{fig:packet-similarity}
\end{figure*}

\noindent\textbf{Task data.}
Table~\ref{tab:training-data} lists the 12 task datasets. After removing exact normalized context overlaps with the test set and invalid duplicate answer texts, the official training pool contains 284,477 questions. A grouped holdout from this pool gives 2,036 development questions, and the remaining 282,441 questions are used for training. The test set contains 55,890 labeled questions from public benchmark splits. These questions are not used for training or model selection. The original option text, option count, and option order are kept. SciQ uses a fixed deterministic option order because its source does not specify one. MuTual uses its labeled development split for testing because its official test labels are not public. We also compute Micro, the accuracy over all 55,890 test questions.

\noindent\textbf{Held-out datasets.}
RACE, DREAM, and QuAIL are used only for testing. RACE and DREAM use their official test sets. QuAIL uses its validation and challenge sets. No held-out passage is an exact match of a Stage 2 training context. Some RACE passages share text with CLOTH training passages, and some DREAM dialogues share text with MuTual training dialogues, because these datasets draw on similar exam material. Removing the affected questions (294 in RACE and 349 in DREAM) changes the accuracy of every method by less than one point.

\noindent\textbf{Token counts.}
Table~\ref{tab:training-data} counts context tokens with the Qwen3.5-0.8B tokenizer, before truncation and without instructions, answer choices, responses, or special tokens. A context shared by several questions is counted once for each question. The reconstruction training set contains 1,133,752,208 context tokens, and the task training set contains 58,554,691. The cross-family pairs use the same texts with their own tokenizers.

\section{Baseline Details}
\label{sec:baseline-details}

\noindent\textbf{Constant packet.}
This control uses the receiver, readers, gates, Stage 1 initialization, data, and training settings of the corresponding \texttt{LatComm} pair: Qwen3.5-4B for the Qwen3.5-0.8B receiver, Qwen3.5-9B for the Gemma-3-1B-pt receiver, and Granite-4.1-8B for the Llama-3.2-1B receiver. The only change is the packet. We encode 2,048 randomly chosen training contexts with the pair's Stage 1 compressors and average their packets. Every training and test question receives this mean packet, so the packet carries no information about the question's context.

\noindent\textbf{Embedding.}
This baseline compresses the context into one vector with a frozen text embedding model. We use F2LLM-v2-14B~\citep{DBLP:journals/corr/abs-2603-19223} and BGE-Multilingual-Gemma2 (9B)~\citep{bge2024gemma2}, each paired with the frozen Qwen3.5-0.8B and Gemma-3-1B-pt receivers. The encoder processes the context as one input, without the question and without chunk pooling. Its last-token representation is L2-normalized, which gives a 5,120-dimensional vector for F2LLM and a 3,584-dimensional vector for BGE. A trainable shared projector maps this vector to one slot. The readers and gates are the same as in \texttt{LatComm}, and they read this slot at every receiver layer: 24 layers for Qwen3.5-0.8B and 26 for Gemma-3-1B-pt. The stored 32-bit vectors take 20\,KiB and 14\,KiB. Each interface is trained from a fresh initialization on the task data, without a reconstruction stage. F2LLM uses a 32,768-token input limit, which truncates no test context. BGE truncates its input to 8,192 tokens. This affects 93 of the 55,890 test questions, which share five documents. For Table~\ref{tab:encoder-choice}, BGE is paired with the Llama-3.2-1B receiver (16 layers). In this run, the BGE vectors are computed in bfloat16 and stored, instead of in 32-bit precision during training. On 15 documents of different lengths, the two computations give a cosine similarity above 0.999.

\noindent\textbf{xRAG.}
We adapt xRAG~\citep{cheng2024xrag}. A frozen SFR-Embedding-Mistral encoder~\citep{sfr2024embeddingmistral} reads the complete context, and a two-layer MLP maps its 4,096-dimensional last-token representation to one soft token placed before the receiver's prompt. Only the MLP is trained, first on reconstruction and then on the task data. Our adaptation differs from the original in three ways: there is no retrieval or document selection, the receiver is Qwen3.5-0.8B, and task training uses the gold option letter without a distillation term. The second xRAG variant replaces the embedding model with the last-token state of Qwen3.5-4B, which has 2,560 dimensions.

\section{Ablation Study}
\label{sec:ablation}

\noindent\textbf{Packet settings.}
Table~\ref{tab:ablation} changes the context encoder or one packet setting at a time. The packet settings use the Qwen3.5-4B encoder and change the number of slots (8, 32, 64), the slot width (128, 256, 512), and the number of layer pairs (24, 12, 6). Reconstruction improves with every increase in capacity: a larger encoder, more slots, wider slots, and more layer pairs all lower the Stage 1 loss and raise the token accuracy. The widest slots give the best reconstruction, with 79.0\% token accuracy, and the highest Task Avg, 66.88. Task Avg changes less than reconstruction. It stays between 63.84 and 66.88 over all configurations, and every configuration is above the best single-vector baseline (61.98). With 8 slots per pair, the packet takes 96\,KiB, one quarter of the default size, and Task Avg is 65.81, against 65.43 for the default packet. With 12 and 6 layer pairs, Task Avg is 64.76 and 63.84. At 96\,KiB, a packet with all layer pairs and 8 slots is therefore more accurate than a packet with 6 layer pairs and 32 slots.

\noindent\textbf{Training loss.}
Figure~\ref{fig:training-loss} shows the training loss of both stages for the same configurations. In Stage 1, the loss decreases throughout training, and the configurations keep the order of Table~\ref{tab:ablation}: more capacity gives a lower reconstruction loss. In Stage 2, the curves are close to each other for the three encoders and for the three slot counts, and the loss is lower with wider slots and with more layer pairs.

\noindent\textbf{Packet controls.}
Table~\ref{tab:packet-controls} tests whether the receiver uses the content of the packet. The same trained readers and gates reach 65.39 Task Avg with the packet of the question's own context, 45.59 with the packet of another context, and 39.60 without a packet. The own packet is the most accurate of the three on all 12 task datasets, and the packet of another context lowers the accuracy by 9.8 to 30.1 points. On the held-out datasets, the own packet reaches 49.82 Held-out Avg, against 36.28 with the packet of another context and 36.41 without a packet. A packet from another context therefore gives no gain on new datasets, and the gain comes from the content of the own packet. The Constant packet control, whose readers and gates are trained with one fixed packet, reaches 60.25 Task Avg and 43.37 Held-out Avg.

\section{Additional Analysis}
\label{sec:analysis}

\noindent\textbf{Seed variation.}
We repeat Stage 2 of the default pair with three seeds. The three runs share the compressors and the packets. Task Avg is 65.43, 65.72, and 65.88, with a mean of 65.68 and a standard deviation of 0.23. The main tables report the first run, which is the lowest of the three, and the gain over the best baseline (61.98) is much larger than the variation between seeds.

\noindent\textbf{Gain source.}
We split the test questions of the 12 task datasets by whether the Qwen3.5-4B encoder answers them correctly when it reads the full text, which holds for 76\% of the questions (Figure~\ref{fig:gain-source}). On these questions, \texttt{LatComm} reaches 72.4\% accuracy, against 68.7\% for the Embedding baseline and 64.5\% for the Constant packet. On the other questions, the three methods are close, with 50.5\%, 48.0\%, and 50.1\%. The gain of \texttt{LatComm} is therefore on the questions that the encoder itself can answer from the context.

\noindent\textbf{Context length.}
Figure~\ref{fig:reuse-cost-length} plots the receiver time per question for each of the 115 QuALITY articles. With text passing, the time grows linearly with the context length, from 265\,ms for the shortest article to 707\,ms for the longest. With a packet, it stays between 130 and 135\,ms for every article, because the receiver always reads the same number of slots.

\noindent\textbf{Gate values.}
We record the gates of the default pair on a sample of 5,386 test questions (Figure~\ref{fig:gate-values}). The mean gate is 0.40, and it differs across layer pairs, from 0.05 at the last pair to 0.59 at pair 18. The slots of one layer pair also differ: 23.4\% of the layer--slot cells have a mean gate below 0.05, and 4.3\% have a mean gate above 0.95. The receiver therefore weights the slots differently at each layer pair, which one gate per layer could not express.

\noindent\textbf{Packet similarity.}
Figure~\ref{fig:packet-similarity} compares the packets of the contexts in the same sample. The slots of one layer pair are nearly orthogonal at the first layer pairs, with a mean cosine similarity of 0.04, and more similar at the last ones, with 0.65. The packets of two contexts from different datasets have a mean cosine similarity of 0.72, so a part of the packet is shared by all contexts, which agrees with the accuracy of the Constant packet control. The other part separates the contexts. Packets from the same dataset are more similar than packets from different datasets at every layer pair. A nearest-mean classifier on the mean slot of a layer pair, fitted on half of the contexts and tested on the rest, identifies the source dataset among 12 with 30\% to 92\% balanced accuracy, against 8.3\% by chance.
\end{document}